\documentclass[pdflatex, sn-basic]{sn-jnl}

\usepackage{graphicx}
\usepackage[english]{babel}
\usepackage[utf8x]{inputenc}
\usepackage{listings}
\usepackage{booktabs}
\usepackage{multirow}
\usepackage{amsmath,amssymb,amsfonts}
\usepackage{amsthm}
\usepackage{mathrsfs}
\usepackage[title]{appendix}
\usepackage{xcolor}
\usepackage{textcomp}
\usepackage{manyfoot}
\usepackage{algorithm}
\usepackage{algorithmicx}
\usepackage{algpseudocode}
\usepackage{cleveref}
\usepackage{subcaption}
\usepackage{todonotes}
\usepackage{placeins}

\DeclareMathOperator*{\argmax}{arg\,max}

\def\modelname{NPLM}
\def\modellongname{Nutrition PPG Language Model}

\title{Wrist Photoplethysmography Predicts Dietary Information}

\author[1]{Kyle Verrier}
\author[1]{Achille Nazaret}
\author[1]{Joseph Futoma}
\author[1]{Andrew C. Miller}
\author[1,2]{Guillermo Sapiro}

\affil[1]{Apple}
\affil[2]{Princeton University}
\date{}

\begin{document}

\abstract{Whether wearable photoplethysmography (PPG) contains dietary information remains unknown. We trained a language model on 1.1M meals to predict meal descriptions from PPG, aligning PPG to text. PPG nontrivially predicts meal content; predictability decreases for PPGs farther from meals. This transfers to dietary tasks: PPG increases AUC by 11\% for intake and satiety across held-out and independent cohorts, with gains robust to text degradation. Wearable PPG may enable passive dietary monitoring.}

\maketitle

\section{Main}

Photoplethysmography (PPG) from consumer wearables have been shown to predict cardiovascular, metabolic, and biological clock information \citep{large-scale-training, miller2025wearable, erturk2025beyond}. Whether PPG signal also captures dietary information—i.e., physiological responses to food intake—remains unknown. We test this hypothesis by aligning PPG representations with meal descriptions using a language model, then measuring whether the alignment carries predictive information about meals. We find that wrist PPG reliably encodes meal-related signals: the signal is strongest near meals, decays with time, persists beyond individual traits, and improves prediction of both caloric intake and satiety in an independent cohort.

Hemodynamic changes following ingestion are well-documented \citep{host1996haemodynamic}, and vascular properties including pulse wave velocity and blood pressure change in response to meals with varying macronutrient composition, though the relationship to peripheral arterial properties measurable by wrist-worn devices remains an active area of investigation \citep{fryer2021central, lithander2013postprandial}. PPG measures microvascular blood-volume dynamics, and PPG-derived features can predict demographics, cardiovascular state, and metabolic health \citep{large-scale-training}, adding value beyond behavioral signals \citep{erturk2025beyond}. However, whether postprandial signals are detectable in wrist-worn PPG, captured opportunistically in real-world conditions with substantial noise and physiological variability, remains unexplored. We hypothesized that if PPG captures postprandial physiology, conditioning on PPG should improve predictions of meal content.

\begin{figure}[p]
     \centering
     \begin{subfigure}[b]{1\textwidth}
         \centering
         \includegraphics[width=\textwidth]{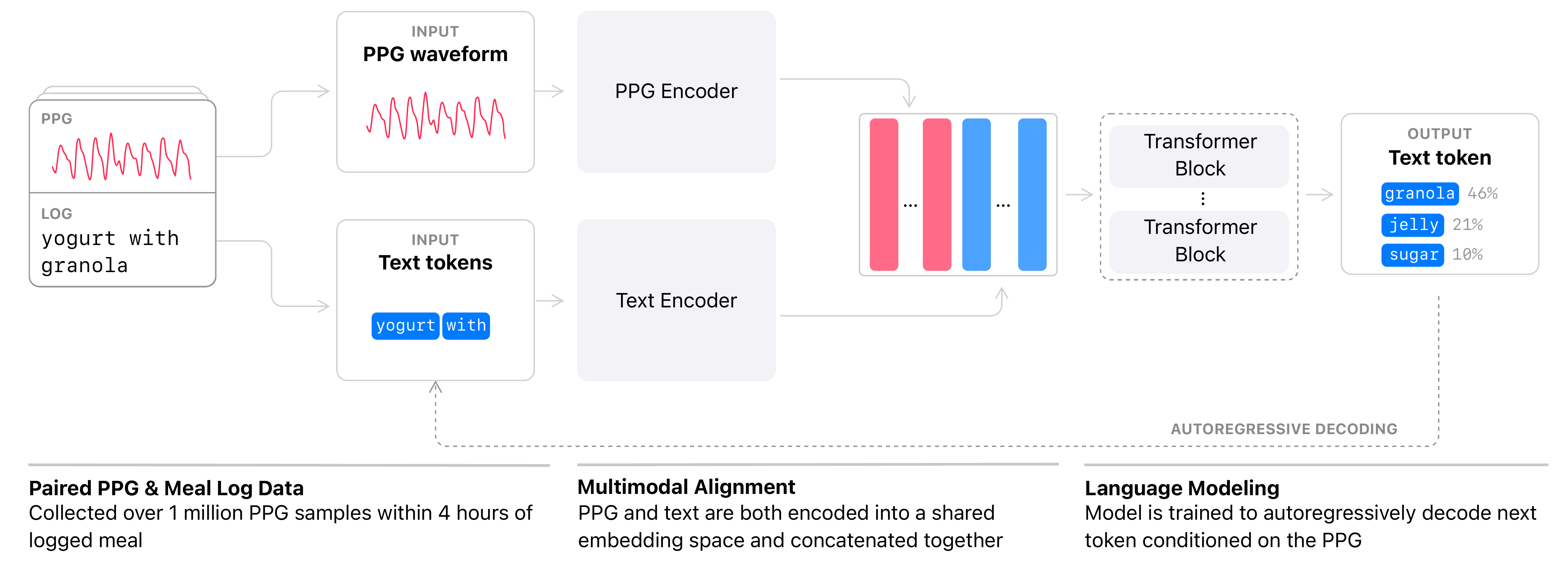}
         \caption{}
         \label{fig:pretraining_a}
     \end{subfigure}
     \vfill
     \begin{subfigure}[b]{0.7\textwidth}
         \centering
         \includegraphics[width=\textwidth]{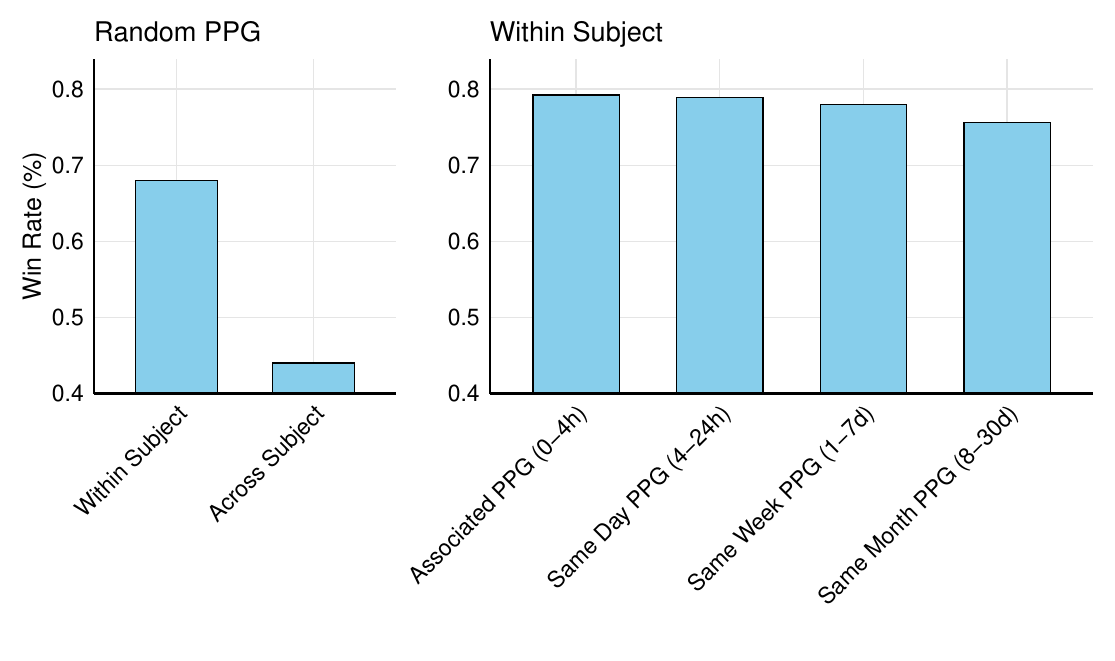}
         \caption{}
         \label{fig:pretraining_b}
     \end{subfigure}
    \caption{\textbf{Alignment of PPG with language reveals a nutritional signal.}
    \textbf{(a)} Diagram of the \modelname{} architecture. Modality-specific encoders project inputs into a shared language embedding space. Alignment is trained by maximizing meal-text likelihood conditioned on near-meal PPG.
    \textbf{(b)} Evidence of Alignment. Left: Changes in win rate when conditioning on time-matched randomly permuted PPG segments \textit{within-subject} versus randomly permuted \textit{across-subject.} Right: Changes in win rate for PPG segments trained at increasing time lags relative to the logged meal within subject; the win rate monotonically decreases.}
    \label{fig:pretraining}
\end{figure}

We introduce \modellongname{} (\modelname{}), which aligns two models: a PPG encoder that learns representations predictive of a variety of health conditions \citep{large-scale-training}, and a language model that learns rich text representations \citep{radford2019language}. Text serves as a flexible target modality that can represent meal information at varying levels of detail. Given a PPG signal, \modelname{} learns to generate meal descriptions likely consumed near PPG collection time (Fig. \ref{fig:pretraining_a}). We define ``near-meal'' as PPG segments from $-$4 to $+$4 hours relative to meal time. This temporal window accommodates noisy timestamps in free-living settings while maximizing data availability. Within this window, we select the PPG segment closest to the meal timestamp. Most selected segments fall within 90 minutes of meal time. This objective induces PPG representations predictive of meal content, exposing nutritional signals within passively collected PPG waveforms.

We used data from two studies. The Apple Heart \& Movement Study (AHMS) was a real-world observational study where participants provided informed consent to share health information for research use \citep{truslow2024understanding}. We included $19{,}340$ subjects with $1{,}122{,}834$ self-reported meal descriptions; AHMS was used to train \modelname{} and evaluate alignment and downstream intake prediction. The Validation Study, an independent controlled study with 140 participants and 720 meals from a dining facility, provided meal descriptions automatically recorded from ordering apps\footnote{Participants ordered meals through restaurant ordering applications, which logged meal contents automatically, reducing self-reporting burden compared to AHMS.} and post-meal appetite ratings; we used this cohort to test the ability of our PPG-to-text alignment to generalize and predict satiety without retraining. PPG was passively collected in both studies. Data filtering details are in Appendix~\ref{sec:data_filtering} and cohort statistics in Table~\ref{tab:summary_statistics}.

Given meal description $X_{\texttt{text}}$ and PPG embedding $X_{\texttt{PPG}, 0\text{-}4\mathrm{h}}$, the modality alignment process maximizes conditional likelihood $p(X_{\texttt{text}} \mid X_{\texttt{PPG}, 0\text{-}4\mathrm{h}})$. High probability indicates PPG contains meal-predictive information. This alignment process is illustrated in \Cref{fig:pretraining_a} and detailed in \Cref{sec:pre-training-alignment}.

To quantify how much meal content information a PPG segment contains, we define \textit{alignment win rate} as the proportion of meals where PPG-conditional likelihood exceeds the regular text-only likelihood given by a capacity-matched text-only baseline (see \Cref{sec:alignment_win_rate}). Using near-meal PPG segments within 4 hours of meal times, we find win rate 0.79 (95\% CI 0.78–0.80), indicating PPG reliably encodes meal-predictive information.

To confirm this signal reflects genuine meal-PPG relationships, we performed two controls (\Cref{fig:pretraining_b}, left). First, permuting time-matched PPG across subjects reduced win rate to 0.43 (95\% CI 0.42–0.44), below chance (50\%), confirming the signal is subject-specific. Second, permuting time-matched PPG within subjects yielded 0.69 (95\% CI 0.67–0.71), indicating persistent individual traits. The gap between within-subject permutation (0.69) and true pairing (0.79) demonstrates that PPG carries meal-specific information beyond persistent traits—the signal depends on \textit{which} meal, not just \textit{who} the person is.

We also tested temporal specificity by aligning meal descriptions to PPG at increasing distances from meals (\Cref{fig:pretraining_b}, right). Win rate decreased monotonically with time, confirming that waveforms closest to ingestion carry most dietary information and ruling out static time-of-day confounds. Some predictive capacity persists even weeks after meals; given substantial meal diversity across individuals (average diversity score 0.54) with weak correlation to performance ($r = -0.09$), these long-term signals likely reflect persistent physiological impacts of dietary patterns rather than memorization.

Having established that PPG contains meal-predictive information, we next tested whether this signal improves prediction of diet-related outcomes. We evaluated two downstream tasks: (1) \textbf{Daily Energy Intake Prediction} on AHMS, predicting whether daily caloric intake exceeds personal mean as a proxy for appetite regulation \citep{drapeau2005appetite, rakha2022insights}; and (2) \textbf{Postprandial Satiety Prediction} on the Validation Study, inferring changes in self-reported hunger-satiety levels following meals.

\begin{figure}[p]
    \centering
    \includegraphics[width=0.9\textwidth]{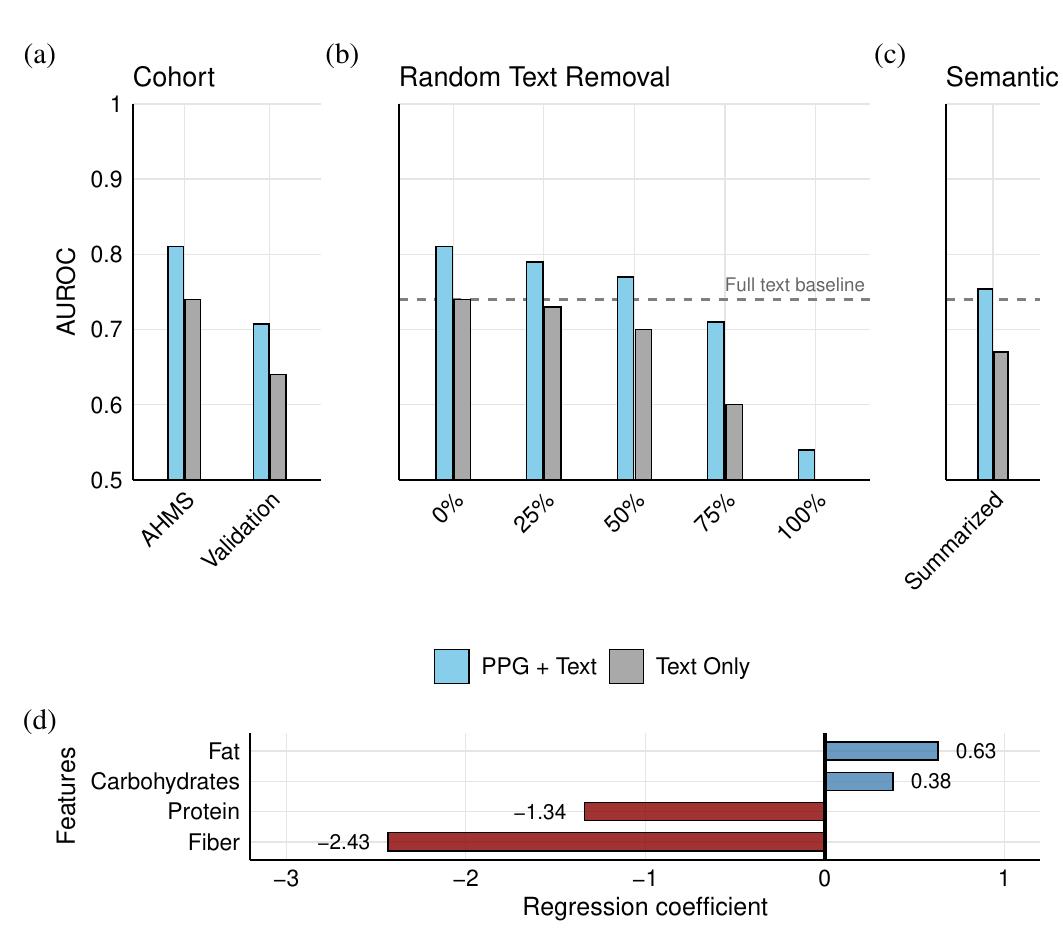}
    \caption{\textbf{\modelname{} predicts dietary outcomes across multiple experimental conditions.}
    \textbf{(a)} \textit{Predicting dietary outcomes across cohorts.} We report the area under the receiver operating characteristic curve (AUC) for daily caloric intake classification (AHMS) and postprandial satiety prediction (Validation Study), comparing multimodal \modelname{} against text-only baselines.
    \textbf{(b)} \textit{Relationship between performance and text length in AHMS.} Removing text lowers average model performance. Text alone outperforms PPG alone, however, PPG with 50\% of the tokens outperforms full text. When removing all text (100\%), the results illustrate that PPG alone has limited information, though it is powerful when added to (simplified) meal description.
    \textbf{(c)} \textit{Performance with semantic summarizations of food logs.} \modelname{} with summarized meal descriptions outperforms the original full-text baseline.
    \textbf{(d)} \textit{Surrogate nutrient model coefficients}.
    }
    \label{fig:model_performance_and_tokens}
\end{figure}

In AHMS, \modelname{} achieved AUC 0.82 (95\% CI 0.814–0.826) for distinguishing days with above-mean caloric intake versus below-mean, compared to text-only baseline 0.74 (95\% CI 0.732–0.748), an 11\% relative improvement (paired $p<0.0001$; Fig.~\ref{fig:model_performance_and_tokens}a). In the Validation Study, without retraining, \modelname{} achieved AUC 0.71 (95\% CI 0.70–0.72) for predicting satiety shifts versus text-only baseline 0.64 (95\% CI 0.63–0.65; paired $p<0.0001$). The consistent improvement across cohorts, outcomes, and text formats suggests the PPG dietary signal generalizes beyond training conditions. Lower absolute performance in the Validation Study likely reflects distributional shift: meal descriptions were 49\% shorter than AHMS, and the outcome (satiety) differs from training (intake).

To evaluate \modelname{}'s resilience to shorter, coarser, or incomplete meal descriptions, we performed ablations on AHMS by randomly removing words or using an LLM to condense descriptions to single keywords (e.g., ``buttermilk pancakes with blueberries'' becomes ``pancakes''; see \Cref{sec:shortened_meal_descriptions}). Text-only AUC declined with text removal, and PPG-only provided modest baseline (Fig.~\ref{fig:model_performance_and_tokens}b-c). In contrast, multimodal \modelname{} retained strong performance: with 50\% of text removed, it still outperformed the full-text-only baseline. Crucially, with LLM-condensed descriptions, \modelname{} achieved AUC 0.75 (95\% CI 0.74–0.76), exceeding full-text-only (0.74). This demonstrates that PPG contributes information beyond what text provides: even minimal meal descriptions combined with PPG outperform detailed text alone, suggesting PPG can compensate for incomplete dietary reporting.

To probe biological plausibility, we trained an interpretable surrogate model \citep{sigut2023interpretable} that predicts \modelname{}'s dietary predictions from nutrient composition when available. Previous work has shown that nutrient composition predicts satiety \citep{drapeau2005appetite, tucker2016postprandial} and language models can estimate nutrient composition from meal descriptions \citep{dhaliwal2025nutribench}; if NPLM captures satiety-relevant signals, its predictions should correlate with nutrients. We observe statistically significant regression coefficients for fat, carbohydrates, fiber and protein (Figure \ref{fig:model_performance_and_tokens}d), consistent with prior work \citep{drapeau2005appetite, tucker2016postprandial, simpson2005obesity, raubenheimer2019protein, hill2000dietary, bornet2007glycaemic}. The relationship between macronutrient composition and satiety is complex and mediated by numerous factors beyond simple nutrient content, including meal timing, energy expenditure, sleep quality, and individual metabolic state. The surrogate model coefficients reflect associations learned by NPLM rather than direct causal mechanisms, and should be interpreted as one simplified view of a multi-factorial process. While we do not directly measure all such factors in this study, prior work has shown that PPG-derived features encode information about behavioral patterns (e.g., activity, sleep), metabolic state, and biological clock \citep{erturk2025beyond, large-scale-training, miller2025wearable}. The observed benefit of incorporating PPG alongside meal text may reflect the model's implicit capture, via a single and comprehensive signal, of such contextual factors that mediate satiety dynamics.

Despite promising findings, our study has limitations. Both cohorts rely on self-reported meal logs and appetite ratings introducing noise and bias \citep{subar2015addressing}; we expect stronger results with controlled measurement. Cohort demographics are skewed: AHMS participants are U.S. Apple Watch users, Validation Study participants from controlled dining, limiting socioeconomic and geographic generalizability. AHMS participants logging meals potentially did so for specific health-related goals (e.g., weight management, diabetes management), which may influence dietary patterns and intake-satiety relationships. Validation Study differed demographically from AHMS, including sex distribution. We did not perform demographic matching between cohorts, as Validation Study was designed to test generalization to an independent population rather than provide a demographically matched replication. While this approach allowed us to assess robustness across different demographic distributions, it limits our ability to make strong inferences about subgroup-specific effects. Methodological limitations include: the $\pm4$ hour temporal window for PPG-to-meal associations is relatively broad and may include confounders such as additional unlogged intake, physical activity, or stress responses; Validation Study assessed satiety exclusively at midday weekday meals, which may not generalize to other meal times given circadian variation in satiety-related hormones; non-caloric intake (water, coffee) not accounted for despite influencing PPG and satiety; behavioral factors like CGM or GLP-1 agonist use not systematically captured. Lack of randomized or longitudinal intervention data precludes causal inference about whether model-driven feedback could improve outcomes. Real-world deployment would require careful attention to data quality, user adherence, and integration with clinical workflows. Despite these limitations, the model provides measurable and actionable information about dietary responses in real-world settings.

By aligning wrist PPG with meal descriptions at population scale and validating in an independent cohort, we provide evidence that vascular pulse signals from consumer wearables encode reproducible signatures of recent intake and satiety dynamics. The signal is temporally specific (strongest near meals), meal-modulated (beyond persistent physiological traits), robust to text degradation, and generalizes across cohorts and outcomes. These findings establish nutrition as a domain where noninvasive, scalable PPG provides physiological insights, motivating prospective studies on metabolic endpoints (weight management, glycemic control) and extension to laboratory biomarkers.

\section{Methods}

\subsection{Study Population and Design}
We use data from two studies: the Apple Heart \& Movement Study (AHMS) and the Validation Study.

\subsubsection*{Apple Heart \& Movement Study}

The Apple Heart \& Movement Study (AHMS) is a research study conducted in partnership with the American Heart Association and Brigham and Women's Hospital from November 14, 2019 to February 28, 2025. The AHMS was designed to study connections between physical activity and cardiovascular health. Participants were using iPhone and Apple Watches, at least 18 years of age (at least 19 years old in Alabama and Nebraska; at least 21 years old in Puerto Rico), who resided in the United States and provided informed consent electronically in the Research app. The study was approved by the Advarra Central Institutional Review Board, and registered to ClinicalTrials.gov (ClinicalTrials.gov Identifier: NCT04198194) \citep{truslow2024understanding}. Data types users consent to sharing, including SensorKit and HealthKit data, are collected via the Research app \citep{apple2025research}.

\subsubsection*{The Validation Study}

The Validation Study was conducted to complement the AHMS by providing structured, controlled meal intake data from a food service facility (restaurant). A study cohort of participants volunteered for a six-week observational period starting on October 11, 2023, during which their daily lunch orders were automatically logged via a customized version of the Research app. Following each lunch meal, participants completed automated postprandial surveys at precise intervals (15, 45, 135, 155, and 195 minutes) to capture changes in perceived fullness and satiety. These surveys were inspired by ecological momentary assessments as defined in \citep{shiffman2008ecological}, triggered \textit{in situ} via a dedicated application immediately after placing an order, and adapted from a validated generalized magnitude scale specifically optimized for iPhone and Apple Watch screens \citep{cardello2005development}. The study provided each participant with a daily voucher covering the cost of a standard lunch. Participants were asked to eat at the facility at least three times per week. This controlled dining setting included standardized nutrition descriptors available to participants when selecting meals, and all meals were consumed at a consistent midday timepoint on weekdays. See \Cref{fig:validation_study_app} for screenshots of the Validation Study app. The Validation Study was designed as an independent test of generalization rather than a demographically matched replication. Demographic differences between cohorts (see Table~\ref{tab:summary_statistics}) allow us to assess model robustness across population subgroups.

The study was approved by the Human Study Review Board at Apple (Study ID: 22-1110-ST2444) (see Supplementary Information \Cref{tab:summary_statistics} for participant characteristics and \Cref{sec:inclusion_exclusion} for inclusion/exclusion criteria), and informed consent was obtained electronically from all subjects. Participants were not compensated for their participation.

\subsection{Data Acquisition and Preprocessing}

\subsubsection*{Photoplethysmography (PPG)}

Apple Watches capture 60-seconds green-light (525 nm) PPG segments during low-motion intervals at 64 Hz or 256 Hz.
We encode each individual PPG segment into 256-dimensional embeddings using the independently pre-trained recently reported PPG encoder \citep{large-scale-training}, which was developed for general-purpose health modeling across multiple research applications. We note that no dietary or satiety information was used during the PPG encoder training.

\subsubsection*{Meal Descriptions}

In the AHMS, some participants record and voluntarily share free-text meal descriptions and associated macronutrient levels using third‑party logging apps (e.g., LoseIt, MyNetDiary, Noom). These apps vary in their user interfaces, prompting strategies, and metadata fields, which may contribute to variability in text length and detail; such variability is expected in real-world settings and this can be considered a strength of the study. Meal logs are transmitted to the AHMS via standard HealthKit integration. In the Validation Study, meal descriptions are automatically populated from the food service facility's ordering system when participants place their orders via the study app, meaning participants did not manually enter free-text descriptions but rather selected items from the available menu. This may result in more standardized but less personalized meal descriptions compared to AHMS; as we will show, our results are robust to these differences, showing the intrinsic value of the proposed model. To ensure data integrity, we remove records with implausible macronutrient values (e.g., negative or extreme outliers) or specific meal descriptions like ``Skipped Meal.'' (See \Cref{sec:data_filtering} for details on the data inclusion procedure.)

\subsubsection*{Sample Splitting}
AHMS data were partitioned 80/20\% into training and held-out test sets at the participant level, with 10\% of the training set reserved for validation and hyperparameter tuning. All the results presented in this paper on AHMS were computed on the held-out test set. The Validation Study data was held out entirely for external testing without any model fine-tuning.

\subsection{Outcome Measures}

For both cohorts, we leveraged learned representations of text and PPG signals to predict satiety‐related outcomes.

\subsubsection*{AHMS Daily Calorie Intake Proxy for Appetite Regulation}
The AHMS study does not include direct satiety assessments. Instead, we look at daily calorie consumption of each subject and aim to predict if subject ate more or less than their individual average each day. We obtain the daily calorie consumption by summing all recorded meal calories for each subject on a given day. To account for inter‐individual differences in eating habits, we z-score each participant's daily total (subtracting their personal mean and dividing by their standard deviation across the study). We interpret higher-than-average intake days as reflecting increased appetite drive or metabolic demand, recognizing this represents an imperfect but scalable proxy for appetite regulation dynamics \citep{drapeau2005appetite, rakha2022insights}.

\subsubsection*{Validation Study and Meal-Specific Satiety}
In contrast, the Validation Study cohort collects direct, meal‐level fullness ratings. Ratings are obtained via a validated generalized magnitude scale adapted for mobile devices \citep{cardello2005development}. For each meal, we define the satiety provided by a meal as the change in self‐reported fullness, subtracting the baseline fullness rating captured immediately after placing the meal order from the average fullness rating reported over the 2–3 hours postprandial window. 

\subsection{Model Architecture and Training}

We aim to predict and understand satiety from textual description of meals and from the associated PPG signals captured around the time of meal consumption.
We leverage the textual comprehension capabilities of pre-trained large language models (LLMs) to understand food descriptions and show how to adapt LLMs to also understand PPGs signals. Obtaining those multimodal capabilities requires an alignment step that we describe next.

\subsubsection*{Multimodal Alignment of PPG and Text}
\label{sec:pre-training-alignment}

We align PPG embeddings with large language model tokens using a learnable linear adapter that transforms a 256-dimensional PPG embedding $x$ into a series of $K$ tokens $f_\theta(x) = [f_{\theta_1}(x), ..., f_{\theta_K}(x)]$ with linear transformations $f_{\theta_1}, ..., f_{\theta_K}$.

We use the pre-trained GPT-2 model \citep{radford2019language} as a backbone text encoder, and pre-trained PPG encoder \citep{large-scale-training}. 
The probability distribution induced by GPT-2 is denoted by $q$, and the PPG encoder is denoted by $e$.
To train the linear adapter, we first consolidate pairs of textual meal description and corresponding PPG segments: (text tokens $X_{\tt{text}}$, PPG signal $X_{\tt{PPG}}$).

We then model the likelihood of a text description given a PPG segment as

\begin{equation}
    p_\theta\bigl(X_{\tt{text}} \mid X_{\tt{PPG}}\bigr)
    = \prod_{i=1}^{L} q \bigl(X_{\tt{text}}^{[i]} \,\Bigm|\,
    f_\theta(e(X_{\tt{PPG}})),\, X_{\tt{text}}^{[<i]}\Bigr).
    \label{eq:linear_alignment}
\end{equation}

Finally, we find the linear adapter $f_\theta$ that maximizes the conditional likelihood of the pairs of text and PPG by solving
\begin{equation}
    \argmax_{\theta} p_\theta\bigl(X_{\tt{text}} \mid X_{\tt{PPG}}\bigr).
    \label{eq:optimization}
\end{equation}

In \Cref{eq:linear_alignment} the only learnable parameter is the linear projection $f_\theta$ from PPG embedding space to LLM token space. The PPG encoder $e$ and the LLM $q$ are frozen. The optimized $\theta$ thus represents the most effective linear adapter for predicting meal description from corresponding PPG signals.

The parameters $\theta$ are optimized using Adam with a learning rate of $3\times10^{-4}$ and batch size of 64. Training iterated for 10{,}000 steps, with early stopping implemented based on validation set performance.

\subsubsection*{Pairing and Temporal Windows}
For each meal description $X_{\tt{text}}$ with timestamp $t_{\text{meal}}$, we construct PPG—text pairs by aggregating passively collected PPG from a prespecified window relative to $t_{\text{meal}}$ and computing an embedding $x = e(X_{\tt{PPG}, w})$.
We define a \emph{near-meal} window $w_{\text{0-4h}} := [-4\text{h}, +4\text{h}]$. To evaluate temporal specificity under distributional shift, we also form three additional windows: $w_{\pm4\text{--}24\text{h}} := [\pm4\text{h}, \pm24\text{h}]$, $w_{\pm1\text{--}7\text{d}} := [\pm1\text{d}, \pm7\text{d}]$, and $w_{\pm8\text{--}30\text{d}} := [\pm8\text{d}, \pm30\text{d}]$. For each window $w$, we construct a dataset
$
\mathcal{D}_{w} = \{ (X_{\tt{text}}, X_{\tt{PPG}, w}) \}
$. To respect distributional differences between windows, we train independent models per window using identical architectures, optimizer, learning rate, batch size, and training budget with early stopping on a subject-disjoint validation set.

\subsubsection*{Satiety Prediction Head}

We use the representations of meals (text and PPG) learned by \modelname{} to predict various outcomes related to meals, e.g., the total calories of the meal or the satiety level provided by the meal. 
To do so, we train linear models on top of the final layer of the LLM. 
Specifically, we apply max pooling across the final-layer representations of all tokens (including both the PPG prefix and meal-text tokens) to obtain a single vector, on which we linearly regress the desired outcome.

\subsubsection*{Model Interpretation}

To interpret the satiety signals captured by \modelname{}, we employed a global surrogate model \citep{sigut2023interpretable}. This method involves training a simpler, more interpretable model that mimics the behavior of the complex \modelname{} used for satiety prediction. Specifically, we used a simple logistic regression model to approximate the satiety outcome predicted by \modelname{}. The surrogate model was trained on the nutrient composition of meals, which is available in the AHMS dataset, and the predictions from \modelname{}. This approach allows us to analyze the coefficients of the surrogate model to understand how different nutrients contribute to satiety predictions.
The objective is to capture the relationships between the
predictors and the target variables in an interpretable manner. We visualized the feature regression coefficients to further interpret the potential internal decision process of \modelname{}.

\subsection{Evaluation}

\subsubsection*{Alignment Win Rate}
\label{sec:alignment_win_rate}

We define the \textit{alignment win rate} as the proportion of meals for which the PPG conditional likelihood exceeds the text-only likelihood for a given window $w$,

\begin{equation}
\texttt{WinRate}(w)=
\mathbb{P}\left[
p\left(X_{\tt{text}} \mid X_{\tt{PPG}, w}\right)
>
p\left(X_{\texttt{text}} \mid X_{\texttt{PPG,cohort}}^{*}\right)
\right],
\label{eq:winrate_methods}
\end{equation}

\noindent where $X_{\texttt{PPG,cohort}}^{*}$ denotes the learned cohort-wide PPG embedding under the fixed-PPG control. This cohort-wide PPG embedding was trained using the same procedure as in \Cref{sec:pre-training-alignment}, but providing all zeros as input to the PPG encoder $e$ in \Cref{eq:linear_alignment}.

We compute this metric over the dataset $\mathcal{D}_{0\text{-}4\mathrm{h}} = \{(X_{\texttt{text}}, X_{\texttt{PPG}, 0\text{-}4\mathrm{h}})\}$, which contains meal texts paired with the closest available in time PPG (``near-meal PPG") within 4 hours of the recorded meal time.

To validate that the alignment captures genuine physiological signals rather than spurious correlations, we evaluate two negative control conditions: (i) an across-subject permutation that shuffles PPG embeddings between different participants, and (ii) a within-subject time-matched negative that swaps meals within the same individual matched by hour-of-day and weekday. We then re-evaluate the win rate under these control conditions to ensure the observed alignment is not due to confounding factors.

\subsubsection*{Binary Satiety Classification}

Model performance was evaluated using the Area Under Receiver Operating Characteristic curve (AUC) metric, where the positive class was defined as meals above a participant's personal mean intake in AHMS and meals with decreased satiety ratings in the Validation Study.

\subsubsection*{Statistical Analyses}

All results reported in this study were evaluated on held-out test subjects who were not included in model fitting. 
Statistical significance of differences in metrics were assessed using paired t-tests. Reported $P$ values are two-sided and a $P$ value $< 0.05$ was regarded as statistically significant. Confidence intervals (95\%) were determined via bootstrapping with 1,000 iterations.

\subsubsection*{Shortened Meal Descriptions with a Large Language Model}
\label{sec:shortened_meal_descriptions}
We prompt a large language model (LLM) to generate a condensed description of each meal log.
We use Gemma 27b \citep{team2025gemma} with a temperature of 0.2. We provide a prompt with a few shot examples to guide the model:

\begin{verbatim}
You are an expert in nutrition analyzing food logs. 
For each meal, I need a simple coarse description that represents its 
nutritional content.
It should ideally be one representative word.

    Example:
    | long_description | representative_word |
    |------------------|---------------------|
    | guacamole keto | guacamole |
    | ranch dressing packet | dressing |
    | orange chicken string bean chicken veggies | chicken |
    | corn nuts ranch | nuts |
    | naked green machine | smoothie |
    | honey salted caramel | candy |
    | wint o green mints | candy |
    | carbcounter | supplement |
    | vegetables mixed steam bag | vegetables |
    | almond butter shake by life cafe | shake |
    | lemon champagne vinaigrette by whats gabby | dressing |
    | chesters puff corn | chips |
    | turkey and cheddar wrap | wrap |
    | club lulu unwich | sandwich |
    | turkey bacon wranch | wrap |
    | serrano pepper u goat cheese burgers | burger |
    | snickerdoole protein cookie g | cookie |
    | vinaigrette aged balsamic | dressing |
    | chicken and herb breakfast sausage | sausage |
    | old fashion | alcohol |
    | buttermilk pancakes with almond milk egg | pancakes |
    | smoked provolone slices. | cheese |
    | deli slices oven roasted | ham |
    

    Structure the response as JSON with keys:
    - long_description: original food description
    - representative_word: single representative word

    The json:
\end{verbatim}

\subsubsection*{Meal Diversity within Subjects}
\label{sec:meal_diversity}

We quantify the diversity of the meals by computing the number of unique meal descriptions logged by each subject divided by the total number of meals logged by that subject for a given time window. We leverage the shortened meal descriptions obtained with the LLM as described in \Cref{sec:shortened_meal_descriptions} to avoid counting minor variations of the same meal as different meals. For example, ``chicken salad'' and ``chicken salad with dressing'' would be counted as the same meal. We compute the meal diversity score for each subject and month (number of unique meals eaten divided by the total number of meals in a given month), and then average across all months to obtain an overall meal diversity score per subject. We visualize the distribution of meal diversity scores across all subjects in \Cref{fig:meal_diversity_histogram}. We also quantify the relationship between meal diversity and likelihood of the meal text conditioned on near-meal PPG using the Pearson correlation coefficient.

\section{Data Availability}

Data are not publicly available. Any request for data will be evaluated and responded to in a manner consistent with the specific language in the study protocol and informed consent form. Requests for data should be addressed to one of the corresponding authors (K.V.).

\section{Code Availability}

Code for all data analyses and statistical modeling was written in Python 3.9. We use pyspark for large-scale data queries, pytorch for SSL pre-training and training prediction models, scipy and numpy for computing confidence intervals (i.e., via the bootstrap), huggingface transformers library for the language modeling, and matplotlib, seaborn. Code sharing will be considered upon request. 

\bibliography{library}

\section*{Acknowledgments}

Authors wish to thank all participants in the Apple Heart \& Movement Study, without whom this research
would not be possible, as well as partners at the American Heart Association and Brigham and Women's
Hospital. We also want to thank all participants in the Validation Study, as well as the food service provider. Finally, we would like to thank the following individuals for thoughtful discussion and guidance in support of this work: Salar Abbaspourazad, Saurabh Arora, Matt Bianchi, Jen Block, Justin Dobson, Cait Dooley, Scott Guelich, Rajiv Kumar, Calum MacRae, Eduardo Martinez Montes, Udhay Nallasamy, Cameron Olsen, Lily Peng, and Prajakta Ranade.

\section*{Author Contributions}
K.V. conceived the research, designed and conducted the Validation Study, developed the model, and conducted analyses. A.N. contributed to model development and analysis. J.F. and A.C.M. provided methodological guidance. G.S. supervised the research. All authors contributed to writing and approved the final manuscript.

\section*{Competing Interests}

The authors declare no competing non-financial interests but the following competing financial interests: All authors are employees of Apple, Inc. and own Apple, Inc. stock. The Apple Heart \& Movement Study received funding from Apple, Inc. and the American Heart Association. The Validation Study was also funded by Apple, Inc.
  
\FloatBarrier

\pagebreak

\begin{appendices}

\section{Supplementary Methods}
\label{sec:supp_methods}

\subsection{Data Filtering Criteria for the Analyses}
\label{sec:data_filtering}
\subsubsection*{Meal-Level Criteria}
To ensure only valid self-reported meals were retained for downstream analysis, we applied the following filters to each logged eating occasion:

\begin{itemize}
  \item \textbf{Caloric bounds:}
    \begin{itemize}
      \item \emph{Minimum energy:} Exclude any meal with reported calories $<50$ kcal.
      \item \emph{Maximum energy:} Exclude any meal with reported calories $>2,500$ kcal.
    \end{itemize}
  \item \textbf{Invalid or non-informative descriptors:} Exclude meals whose description exactly matches any of: ``Skipped,'' ``Custom Generic Meal,'' or ``Nothing.''
  \item \textbf{Incomplete nutrient information:} Exclude any meal lacking a calorie value.
  \item \textbf{Duplicate and timestamp errors:} If two entries for the same participant share identical timestamps and descriptions, retain only the first.
\end{itemize}

\subsubsection*{Subject-Level Criteria}
\begin{itemize}
    \item Contains at least 5 days with at least one meal logged per day.
\end{itemize}

\noindent
For all analyses and figures, participants are weighted equally regardless of the number of meals logged.

\subsection{Participant Inclusion and Exclusion Criteria}
\label{sec:inclusion_exclusion}
\subsubsection*{AHMS}
General Inclusion Criteria:
\begin{itemize}
    \item Use both an iPhone and an Apple Watch.
    \item Be at least 18 years old ($\geq 19$ in Alabama and Nebraska; $\geq 21$ in Puerto Rico).
    \item Reside in the United States.
    \item Provide electronic informed consent via the Research app.
\end{itemize}

\noindent
The AHMS applied minimal exclusion criteria at enrollment. For the current analysis, we did not retrospectively exclude participants based on medical conditions (e.g., type 1 or type 2 diabetes, pregnancy, use of GLP-1 agonists) that were excluded in the Validation Study. This difference in cohort composition should be considered when comparing results across studies, while it can also be considered a demonstration of the generality of the findings here reported.

\subsubsection*{Validation Study}

Participants were screened via questionnaire and medical history with inclusion criteria:

\begin{itemize}
      \item Be at least 18 years old.
      \item Be eligible to participate in the controlled dining study and have access to the participating food service facility.
      \item Be willing to eat at the food service facility three times per week during the Study.
      \item Have an iPhone (8 or higher) running iOS 15 and Apple Watch running WatchOS 8, both of which you have been using as carry devices for at least 3 months.
      \item Be willing to wear your Apple Watch daily and for at least 8 hours per day.
      \item Provide electronic informed consent via the Validation Study app.

    \end{itemize}
General exclusion criteria:
    \begin{itemize}
      \item Are under care for chronic medical conditions including eating disorders, type 1 diabetes, or type 2 diabetes.
      \item Cannot safely eat meals offered at the food service facility, e.g., you have an allergy to ingredients used or you are unwilling to consume these foods.
      \item Have had gastrointestinal surgery within the past year.
      \item Are pregnant.
      \item Have had a heart attack (myocardial infarction), stroke/transient ischemic attack (TIA), or major surgery in the last two months of the study.
      \item Expecting an event during your participation in the Study that will cause substantially different behavior from your usual behavior, e.g., surgery or going on a multi-week vacation.
    \end{itemize}

\newpage

\section{Supplementary Tables}
\begin{table}[ht]
    \centering
    \small
    \begin{tabular}{@{}lrr@{}}
        \toprule
        \textbf{Variable} & \textbf{AHMS Cohort} & \textbf{Validation Study Cohort} \\
        \midrule
        Meals & 1,122,834 & 720 \\
        Subjects & 19,340 & 140 \\
        \addlinespace
        \multicolumn{3}{l}{\textit{Sex, n} (\%)} \\
        \quad Male & 3,966 (59.5\%) & 47 (33.6\%) \\
        \quad Female & 2,379 (35.7\%) & 89 (63.6\%) \\
        \addlinespace
        \multicolumn{3}{l}{\textit{Race, n} (\%)} \\
        \quad White & 4,885 (73.3\%) & 63 (45.0\%) \\
        \quad Asian & 399 (6.0\%) & 66 (47.1\%) \\
        \quad Native Hawaiian/Pacific Islander & 9 (0.1\%) & 1 (0.7\%) \\
        \quad Black/African American & 306 (4.6\%) & 4 (2.9\%) \\
        \quad American Indian/Alaska Native & 34 (0.5\%) & 1 (0.7\%) \\
        \quad Other & 410 (6.2\%) & 2 (1.4\%) \\
        \quad Not reported & 119 (1.8\%) & 3 (2.1\%) \\
        \addlinespace
        Age, median [years] (min, max) & 45 (18, 96) & 40 (25, 65) \\
        BMI, median & 27.6 & 23.4 \\
        \bottomrule
    \end{tabular}
    \caption{Summary statistics of the AHMS and Validation Study cohorts.}
        \label{tab:summary_statistics}

  \end{table}

\newpage
  
\section{Supplementary Figures}

\begin{figure}[h]
    \centering
    \includegraphics[width=0.8\textwidth]{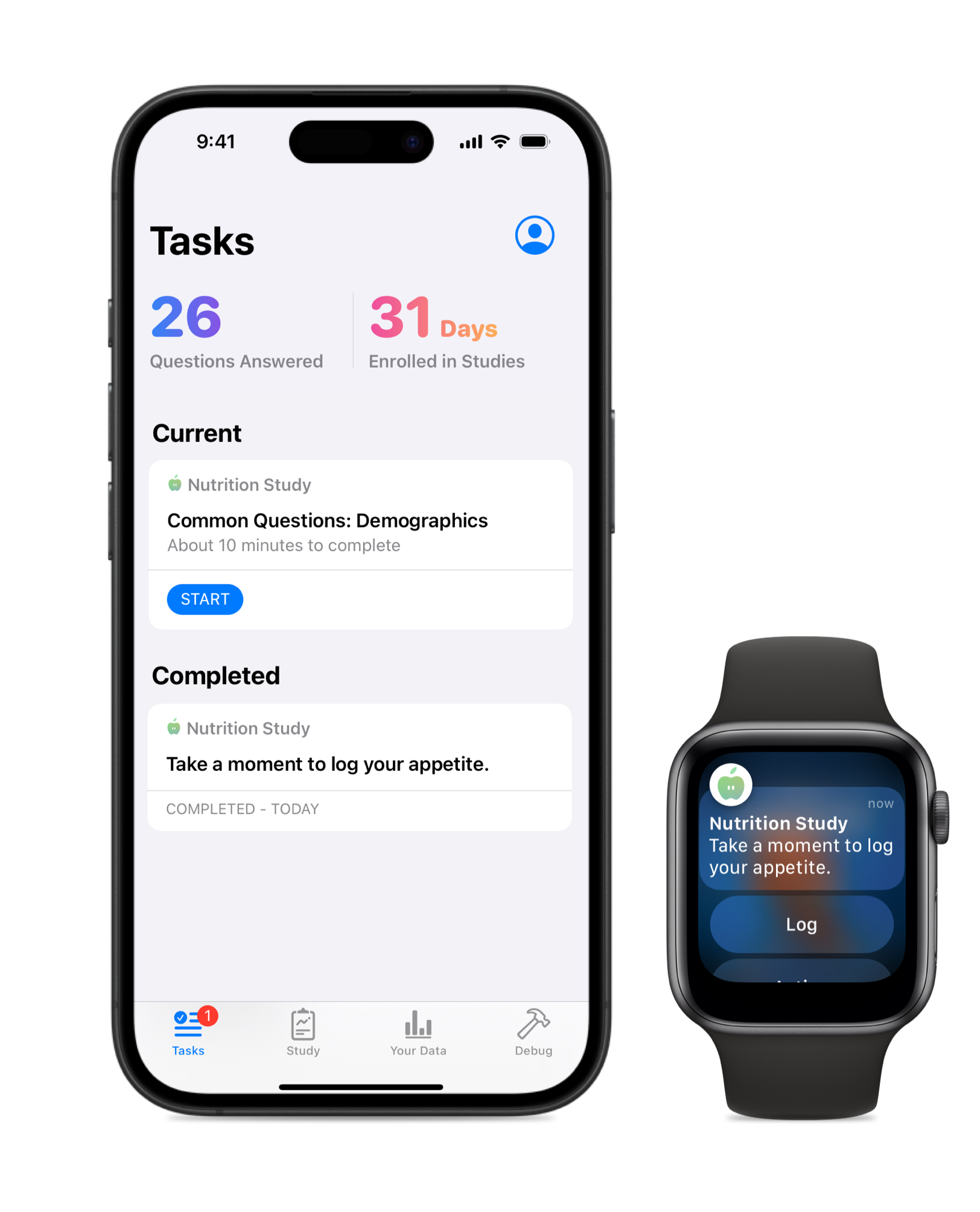}
    \caption{Screenshots of the Validation Study app. The app allows participants to log their appetite and access study information.}
    \label{fig:validation_study_app}
\end{figure}

\begin{figure}[p]
    \centering
    \includegraphics[width=0.8\textwidth]{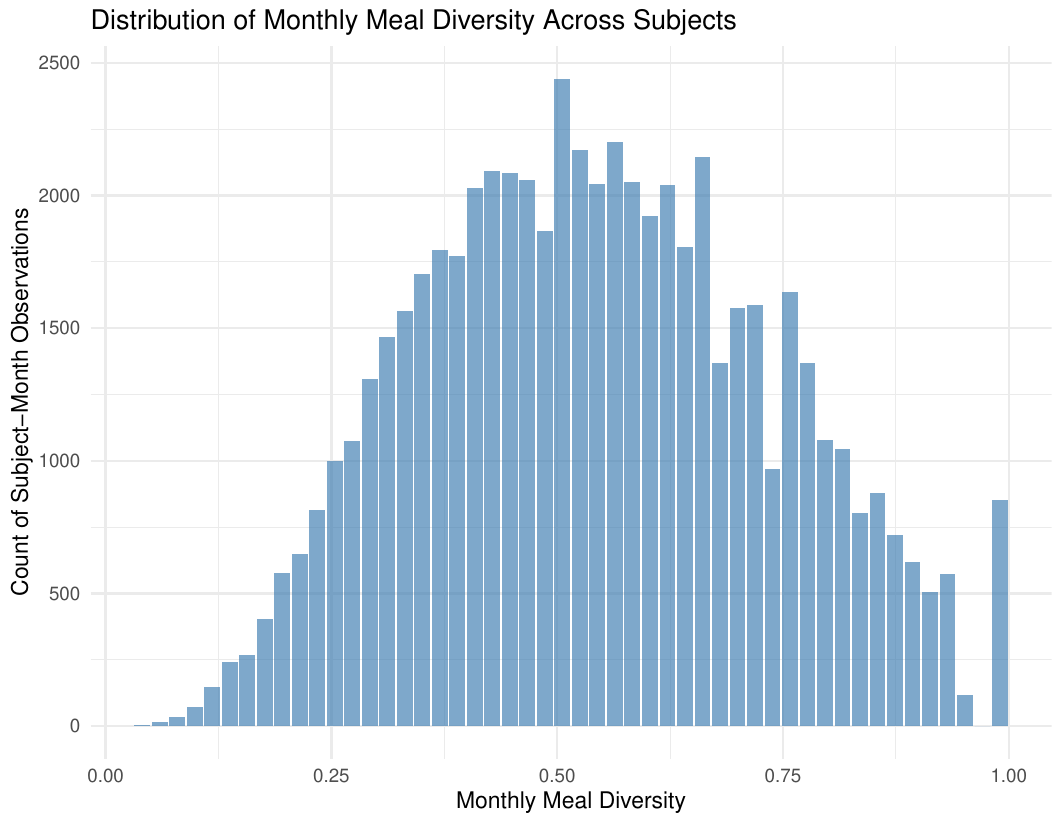}
    \caption{Histogram of meal diversity scores across all individuals in the AHMS cohort. The meal diversity score is defined as the number of unique meals eaten divided by the total number of meals in a given month. The average meal diversity score across all individuals is 0.54. Each participant is weighted equally in this distribution, regardless of the total number of meals logged (inclusion criterion: at least 5 days with at least one meal logged per day).}
    \label{fig:meal_diversity_histogram}
\end{figure}

\end{appendices}

\end{document}